\documentclass[sigconf]{acmart}

\usepackage{multirow}  
\usepackage{graphicx}
\usepackage{fancyhdr}
\pdfoutput=1

\AtBeginDocument{%
  }

\begin{document}

\title{Geometric Feature Enhanced Knowledge Graph Embedding and Spatial Reasoning}

\author{Lei Hu}
\affiliation{%
  \institution{State Key Laboratory of Resources and Environmental Information System, Institute of Geographic Sciences and Natural Resources Research, CAS}
  \institution{University of Chinese Academy of Sciences}
  \city{Beijing}
  \country{China}
}
\affiliation{%
  \institution{School of Geographical Sciences and Urban Planning, Arizona State University}
  \city{Tempe}
  \state{AZ}
  \country{USA}
}
\email{hul@lreis.ac.cn}

\author{Wenwen Li}
\authornote{Corresponding Author.}
\affiliation{%
  \institution{School of Geographical Sciences and Urban Planning, Arizona State University}
  \city{Tempe}
  \state{AZ}
  \country{USA}
}
\email{wenwen@asu.edu}

\author{Yunqiang Zhu}
\affiliation{%
  \institution{State Key Laboratory of Resources and Environmental Information System, Institute of Geographic Sciences and Natural Resources Research, CAS}
  \institution{University of Chinese Academy of Sciences}
  \city{Beijing}
  \country{China}
}
\email{zhuyq@lreis.ac.cn}

\begin{abstract}

Geospatial Knowledge Graphs (GeoKGs) model geoentities (e.g., places and natural features) and spatial relationships in an interconnected manner, providing strong knowledge support for geographic applications, including data retrieval, question-answering, and spatial reasoning. However, existing methods for mining and reasoning from GeoKGs, such as popular knowledge graph embedding (KGE) techniques, lack geographic awareness. This study aims to enhance general-purpose KGE by developing new strategies and integrating geometric features of spatial relations, including topology, direction, and distance, to infuse the embedding process with geographic intuition. The new model is tested on downstream link prediction tasks, and the results show that the inclusion of geometric features, particularly topology and direction, improves prediction accuracy for both geoentities and spatial relations. Our research offers new perspectives for integrating spatial concepts and principles into the GeoKG mining process, providing customized GeoAI solutions for geospatial challenges.

\end{abstract}




\settopmatter{printacmref=false} 
\setcopyright{none}
\renewcommand\footnotetextcopyrightpermission[1]{} 
\pagestyle{fancy} 
\fancyhf{} 

\fancyhead[L]{\footnotesize Geometric Feature Enhanced Knowledge Graph Embedding and Spatial Reasoning}
\fancyhead[R]{\footnotesize Hu et al.}
\maketitle
\section{Introduction}

Geospatial Knowledge Graphs (GeoKGs) as a semantic network, can link multi-source geospatial data, which is essential for geospatial tasks including near real-time disaster response, urban planning, and various applications scenarios involving spatial retrieval, question-answering and reasoning \cite{ijgi12030112}. While current, rapidly evolving Large Language Models (LLMs) excel in language understanding and can perform similar tasks, they lack geographic awareness and are prone to generating inaccurate answers due to their generative nature \cite{roberts2023}. GeoKGs built with geospatial data from authoritative sources, present a trustworthy GeoAI solution for geospatial knowledge users, making them ideal for high-stakes spatial decision-making \cite{li2022performance, li2020geoai}.

KGs typically face issues of incompleteness \cite{KGCompleteness}, and GeoKGs are no exception. Automated KG completion has become a desirable solution for enriching the information and knowledge contained within the KG, and link prediction is a typical task to achieve this goal by predicting missing triplets based on existing entities and relationships in the KG. In recent years, knowledge graph embedding (KGE) has proven to be the state-of-the-art method for link prediction \cite{KGE}. It transforms a large number of entities and relations represented symbolically in the KG into numeric vectors in a low-dimensional space for efficient computation and reasoning while preserving the structure of the KG.

Despite the rapid progress in KGE, challenges remain in applying general-purpose embedding models on geographic entities and their relations. Geographic entities (e.g., a place or a road) are usually contained in a geometric configuration represented by points, lines, or polygons, and the relationships between them typically have spatial properties. The current training paradigm of KGE, relying solely on the semantic structure of triplets, fails to recognize important geometric features, such as directions, distance, and topology. As a result, the selection of candidates in downstream link prediction tasks may not align with geographic intuition, negatively affecting the prediction accuracy. For example, relations that should be identified as ``cross,'' ``span,'' or ``flow through'' may instead be predicted as ``far.''

To address this challenge, we propose a method to inject geographic geometric features into the KGE process. By narrowing the distance between spatial relation terms and their corresponding geometric features in the embedding space, terms that describe the same geometric configuration but with different linguistic conventions will become closer to each other, as demonstrated in Figure 1. This enhancement boosts the validity of the KGE applied in GeoKGs. 

The remainder of the paper is organized as follows: Section 2 describes our GeoKG construction and the geometric feature-enhanced KGE process. Section 3 presents and analyzes the results. In Section 4, we conclude our work and discuss its implications and future research directions.

\section{Data and Methodology}

\subsection{Dataset construction}

Before injecting geographic geometric features into the GeoKG embedding process, we first constructed a GeoKG containing geographic entities and their spatial relations by mining relevant information from Wikipedia pages covering three states: Arizona, Utah, and Colorado. GPT-4 is utilized in the automated construction of GeoKG as described in \cite{srkge}, and based on this, we selected 78 spatial relation terms and their associated triplets for the experiment. We then obtained the footprints of geographic entities from OpenStreetMap \cite{OpenStreetMap}, and computed the geometric features including topology, direction, and distance between each pair of geographic entities in PostgreSQL. These geometric features of spatial relations are summarized in Table 1. Note that we compute distance and direction based on the centroid of the footprints, with distance calculated using an isometric projection.

\subsection{Model construction}

We explore the role of geometric feature incorporation using the well-performing baseline KGE method, the Hierarchy-Aware Knowledge Graph Embedding (HAKE) model, which has shown strong performance due to its multilevel portrayal of entities in the KG \cite{zhang2022HAKE}. We denote the vectors of spatial relations terms in GeoKG as $\mathbf{R}_{{t}}$ and actual geometric features, calculated based on the footprints, as $\mathbf{R}_{{g}}$.

\begin{table}[htb]
    \caption{Summary of Dataset}
    \label{tab:geographic_data}
    \begin{tabular}{lcc}
    \toprule
    Category & Number & Comments \\
    \midrule
    Geo-Entities & 3,180 & - \\
    Spatial relation terms & 78 & - \\
    Geo-Triplets & 4,613 & - \\
    Topology & 58 & 9-Intersection model \\
    Direction & 8 & 8-Point compass \\
    Distance & 20 & Natural breaks (Jenks) \\
    \bottomrule
    \end{tabular}
\end{table}

In HAKE, the concentric circle structure in the polar coordinate system is used to distinguish entities at the same level (different phases \(p\) within one circle) and entities at different levels (different moduli \(m\) across circles) within a KG. Based on this, the triplet $(h, \mathbf{R}_{{t}}, t)$ is modeled in two parts. For the phase part, the relation $\mathbf{R}_{{t,p}}$ is modeled as a phase transformation from the head entity \(h_p\) to the tail entity \(t_p\) through $t_p = (h_p + \mathbf{R}_{{t,p}})\mod2\pi$, where \( h_p, \mathbf{R}_{{t,p}}, t_p \in [0, 2\pi)^k \). For the modulus part, the relation $\mathbf{R}_{{t,m}}$ is modeled as a scaling transformation through $t_m = h_m \circ \mathbf{R}_{{t,m}}$, where \( h_m, t_m \in \mathbb{R}^k \) and \( \mathbf{R}_{{t,m}} \in \mathbb{R}_+^k \). The KGE process constructs a distance function to ensure these two parts hold true simultaneously. 

Building on this, we developed a new distance function that can narrow the distance between $\mathbf{R}_{{t}}$ and $\mathbf{R}_{{g}}$, as shown in Equation 1, with $\lambda \in \mathbb{R}$ as a model learning parameter. To ease computation, we divided the vectors $\mathbf{R}_{{g}}$ into two parts, $\mathbf{R}_{{g,p}}$ and $\mathbf{R}_{{g,m}}$. This ensures that spatial relation terms sharing the same geometric features are positioned closely within the embedding space, thereby enhancing the effectiveness of KGE for geospatial data.

\begin{equation}
\left\{
\begin{aligned}
&d_{\mathbf{R}_{{t}}}(\mathbf{R}_{{g}}) = d_{\mathbf{R}_{{t,m}}}(\mathbf{R}_{{g,m}}) + \lambda d_{\mathbf{R}_{{t,p}}}(\mathbf{R}_{{g,p}}) \\
&d_{\mathbf{R}_{{t,m}}}(\mathbf{R}_{{g,m}}) = \|\mathbf{R}_{{t,m}} - \mathbf{R}_{{g,m}}\|_2 \\
&d_{\mathbf{R}_{{t,p}}}(\mathbf{R}_{{g,p}}) = \|\sin ((\mathbf{R}_{{t,p}} - \mathbf{R}_{{g,p}})/2)\|_1
\end{aligned}
\right.
\end{equation}

We adopt the negative sampling loss function \cite{mikolov2013} to optimize the training for both the HAKE model and the geometric feature enhancement process, the latter as detailed in Equation 2, where \(\sigma\) represents the sigmoid function and \(\gamma>0\) is the fixed margin. We construct negative cases by randomly replacing the geometric features associated with spatial relation terms and sampling them from the distribution $p(\mathbf{R}_{{t}}, \mathbf{R}_{{g,j}}' \mid \{ (\mathbf{R}_{{t,i}}, \mathbf{R}_{{g,i}}) \})$, which is known as the self-adversarial negative sampling strategy \cite{rotate}. Throughout the optimization process, our aim is to minimize the distance function $d_{\mathbf{R}_{{t}}}(\mathbf{R}_{{g}})$ in positive cases and maximize it in negative cases, thereby establishing a mapping between spatial relation terms and their corresponding geometric features.

\begin{equation}
L = -\log\sigma \left( \gamma - d_{\mathbf{R}_{{t}}}(\mathbf{R}_{{g}}) \right) - \sum_{i=1}^n p(\mathbf{R}_{{t}}, \mathbf{R}_{{g,i}}') \log\sigma \left( d_{\mathbf{R}_{{t}}}(\mathbf{R}_{{g,i}}') - \gamma \right)
\end{equation}

\begin{figure}
  \centering
  \includegraphics[width=\linewidth]{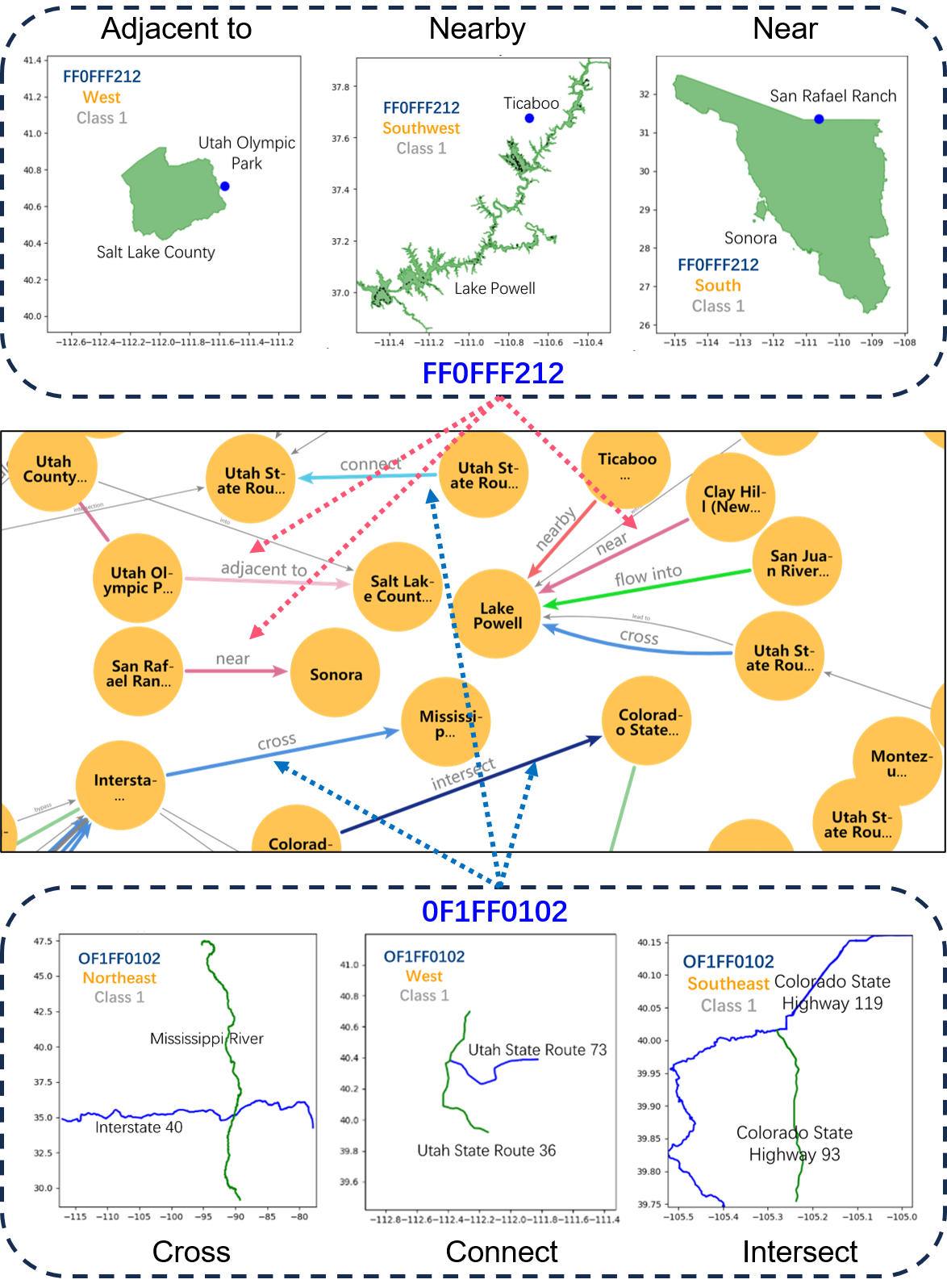}
  \caption{Schematic of the GeoKG knowledge linking process enhanced by geometric features}
  \Description{}
\end{figure}
\subsection{Experiment settings}

We allocate the training, validation, and test sets in an approximate 87:3:10 ratio and utilize the Adam algorithm \cite{kingma2017adam} to optimize the loss function. We set the dimensions of the phase and modulus embeddings all to 200, with \(\gamma = 0.01\), \({lr} = 0.01\), and \({neg\_rate} = 5\). All results were trained for 1000 epochs. We set link prediction as the downstream task to evaluate the model performance, where for any two elements of a triplet $(h, \mathbf{R}_{{t}}, t)$, we compute the HAKE distance function and rank the results in ascending order to predict the third element; a smaller distance function indicates a higher ranking of the prediction result. In experiments, we employ a filtered setting to exclude training triplets from the prediction outcomes, ensuring an unbiased evaluation. We apply the Mean Reciprocal Rank (MRR) and Hits@N metrics for assessment, where a higher MRR signifies better ranking performance and a higher Hits@N indicates a greater proportion of correct predictions among the top N predicted results.

\section{Result}

During the KGE process, we incorporated the geometric features of spatial relations, including topology, direction, and distance, along with triplet information based on the HAKE model (*). As shown in Table 2, the results demonstrated that the geometric features enhance the link prediction performance for the GeoKG.

Specifically, for predicting geographic entities, both topology and direction are important as single features. When considering composite features, the simultaneous inclusion of almost any two features typically results in varying degrees of improvement compared to adding a single feature alone. When all three features are included, the geographic entity prediction achieves its best performance. For predicting spatial relation terms, the topology feature is the most important, a finding that is consistent with previous spatial cognitive research \cite{Egenhofer1998, shariff1998natural}, but adding direction and distance to this does not seem to further enhance the relation prediction. Overall, the simultaneous inclusion of topology, direction, and distance features enhances the KGE model, with topology and direction being the most important features.
To explore the generalizability of the present results, we manually discriminate the main geometric features of the spatial relation terms in the experimental dataset, and a term may be associated with more than one feature, e.g., we manually consider that the term ``surround'' is related to both topology and distance. Statistically, the number of triplets related to topology, direction, and distance in our GeoKG dataset is in the ratio of 69:17:14. The prevalence of spatial relation terms with topology features in the dataset may also explain the primary enhancement of KGE by topology information. If the GeoKG primarily describes relations of distance or direction, the enhancement effects on the model by a single feature may vary.

\begin{table}
  \centering
  \caption{Link Prediction Performance Comparison}
  \label{tab:model_performance}
  \begin{tabular}{c|c|@{\hspace{1pt}}c|@{\hspace{1pt}}c|@{\hspace{1pt}}c|@{\hspace{1pt}}c}
    \toprule
    \textbf{Model} & \textbf{MRR} & \textbf{Hits@1} & \textbf{Hits@3} & \textbf{Hits@5} & \textbf{Hits@10} \\
    \midrule
    \multicolumn{6}{c}{\textbf{Geographic Entity Prediction}} \\
    \midrule
    HAKE & 0.191 & 0.134 & 0.209 & 0.238 & 0.290 \\
    Topo* & 0.205 & 0.151 & 0.214 & 0.249 & 0.310 \\
    Dir* & 0.206 & 0.152 & 0.214 & 0.248 & 0.306 \\
    Dis* & 0.202 & 0.148 & 0.212 & 0.251 & 0.309 \\
    Topo+Dir* & 0.213 & 0.158 & 0.220 & 0.256 & \textbf{0.321} \\
    Topo+Dis* & 0.204 & 0.147 & 0.223 & 0.250 & 0.316 \\
    Dir+Dis* & 0.208 & 0.151 & 0.219 & 0.259 & 0.317 \\
    Topo+Dir+Dis* & \textbf{0.215} & \textbf{0.165} & \textbf{0.221} & \textbf{0.259} & 0.311 \\
    \midrule
    \multicolumn{6}{c}{\textbf{Spatial Relation Term Prediction}} \\
    \midrule
    HAKE* & 0.184 & 0.090 & 0.184 & 0.246 & 0.366 \\
    Topo* & \textbf{0.204} & \textbf{0.106} & 0.198 & \textbf{0.278} & \textbf{0.394} \\
    Dir* & 0.194 & 0.098 & 0.192 & 0.250 & 0.370 \\
    Dis* & 0.187 & 0.090 & 0.176 & 0.256 & 0.380 \\
    Topo+Dir* & 0.198 & 0.100 & 0.196 & 0.274 & 0.390 \\
    Topo+Dis* & 0.195 & 0.096 & 0.192 & 0.270 & 0.390 \\
    Dir+Dis* & 0.182 & 0.086 & 0.176 & 0.254 & 0.352 \\
    Topo+Dir+Dis* & 0.190 & 0.086 & \textbf{0.202} & 0.276 & 0.370 \\
    \midrule
    \multicolumn{6}{c}{\textbf{Overall}} \\
    \midrule
    HAKE & 0.189 & 0.119 & 0.201 & 0.241 & 0.315 \\
    Topo* & 0.204 & 0.136 & 0.209 & 0.259 & 0.338 \\
    Dir* & 0.202 & 0.134 & 0.207 & 0.249 & 0.327 \\
    Dis* & 0.197 & 0.129 & 0.200 & 0.253 & 0.333 \\
    Topo+Dir* & \textbf{0.208} & \textbf{0.139} & 0.212 & 0.262 & \textbf{0.344} \\
    Topo+Dis* & 0.201 & 0.130 & 0.206 & 0.257 & 0.341 \\
    Dir+Dis* & 0.199 & 0.129 & 0.205 & 0.257 & 0.329 \\
    Topo+Dir+Dis* & 0.207 & \textbf{0.139} & \textbf{0.215} & \textbf{0.265} & 0.331 \\
    \bottomrule
  \end{tabular}
\end{table}

\begin{table*}[ht]
    \centering
    \caption{Top 5 head geographic entity predictions.}
    \begin{tabular}{|c|c|c|c|}
    \hline
    \textbf{Relation (r) \& Tail (t)} & \textbf{Rank} & \textbf{HAKE} & \textbf{Topo+Dir+Dis*} \\
    \hline
    \multicolumn{1}{|c|}{\multirow{5}{*}{\parbox{3cm}{\textbf{r: adjacent to}\\ \textbf{t: Phoenix, Arizona}}}} 
     & 1 & Avondale, Arizona & Avondale, Arizona \\
     & 2 & Arizona State Route 85 & Arizona State Route 85 \\
     & 3 & U.S. Route 80 in Arizona & Salt River (Arizona) \\
     & 4 & \textbf{Douglas, Arizona} & Arizona State Route 202 \\
     & 5 & Salt River (Arizona) & Interstate 10 in Arizona \\
    \hline
    \multirow{5}{*}{\parbox{3cm}{\textbf{r: proximity}\\ \textbf{t: Phoenix, Arizona}}} 
     & 1 & \textbf{Interstate 19} & Avondale, Arizona \\
     & 2 & Arizona State Route 85 & Arizona State Route 85 \\
     & 3 & Salt River (Arizona) & Central Arizona Project (a canal) \\
     & 4 & Interstate 10 in Arizona & Casa Grande, Arizona \\
     & 5 & Arizona State Route 87 & Arizona State Route 87 \\
    \hline
    \end{tabular}
    \caption*{\normalfont \textit{Note: Inappropriate predictions are in bold.}}
\end{table*}

As demonstrated in Table 3, we predicted the head geographic entities in triplets (?, adjacent to, Phoenix (AZ)) and (?, proximity, Phoenix (AZ)) using both the HAKE model and the geometric feature-enhanced HAKE (Topo+Dir+Dis*). The inclusion of geometric features helps exclude inappropriate entities, such as ``Douglas, Arizona'' and ``Interstate 19,'' from the predictions made by the HAKE model under specified spatial relations. In fact, they are not ``adjacent to'' or in ``proximity'' to Phoenix, Arizona. Additionally, the geometric feature-enhanced model facilitates the prediction of new relevant geographic entities, including ``Arizona State Route 202'' and ``Interstate 10 in Arizona'' for the relation ``adjacent to,'' and ``Avondale, Arizona,'' ``Central Arizona Project,'' and the relatively nearby ``Casa Grande, Arizona'' for the relation ``proximity,'' thereby enhancing the geographical accuracy of link prediction.



\section{Conclusion and discussion}
In this research, we propose a geometric feature-enhanced GeoKG embedding method for spatial reasoning. By injecting geometric features of spatial relations into the KGE process, our results show an enhancement effect on the prediction of both geographic entities and spatial relation terms in the link prediction task, leading to outcomes that are more consistent with geographic expectations.

Building upon these promising results, our research holds important implications for GeoAI research. First, the enhancement of KGE with geometric features automatically establishes mappings between spatial relations and geometric configurations. This development is valuable for the study of spatial relations in GIScience \cite{mark2012cognitive}. This area of research has traditionally relied on labor-intensive cognitive experiments to link terms with spatial configurations, a method constrained by small-scale studies. In comparison, our model incorporates the inherent diversity of spatial relations, especially those described in natural language, into the GeoKG modeling and prediction. This approach offers a more efficient method for investigating how individuals cognitively engage with geographic spaces. Moreover, with its highly extendable design, our method of integrating geometric features is equally applicable to other KGE models while greatly enhancing their ``Geo'' adaptability.

In the future, we will further explore the scalability of our approach to larger datasets and its applicability across different geospatial contexts. Furthermore, we aim to enhance the robustness of the model by refining the geometric features of spatial relations with more complex metric indices, thereby achieving greater improvements in effective knowledge mining and spatial reasoning of GeoKG.

\begin{acks}
This work is supported in part by the US National Science Foundation (Grant No. 1853864, 2230034, and 2303748), the National Key Research and Development Program of China (Grant No. 2022YFB3904200, 2022YFF0711601, 2021YFB00903), and the Key Project of Innovation LREIS (Grant No. KPI009).
\end{acks}

\bibliographystyle{ACM-Reference-Format}
\bibliography{sample-base}
\end{document}